\documentclass[11pt]{article}
\usepackage{times}
\usepackage{latexsym}
\usepackage{named}
\newenvironment{myitemize}{
\vspace{-0.3\baselineskip}
\begin{itemize}
  \setlength{\topsep}{0pt}
  \setlength{\itemsep}{1pt}
  \setlength{\parskip}{0pt}
  \setlength{\parsep}{0pt}
  \setlength{\partopsep}{0pt}
}{
\end{itemize}
\vspace{-0.2\baselineskip}}
\makeatletter
\def\leftcite{(}\def\rightcite{)}
\def\@biblabel#1{\def\citeauthoryear##1##2{##1, ##2}(#1)\hfill}
\makeatother
\makeatletter
\newcommand\figcaption{\def\@captype{figure}\caption}
\newcommand\tabcaption{\def\@captype{table}\caption}
\makeatother
\makeatletter
\def\maketitle{
\null
\thispagestyle{empty}
\vfill
\begin{center}\leavevmode
  \normalfont
  {\LARGE \@title\par}
  \vskip 2cm
  {\Large \@author\par}
  \vskip 4cm
  {\Large \@date\par}
\end{center}
\vfill
\null
\cleardoublepage}
\makeatother
\title{Analogy Perception Applied to Seven Tests of \\
Word Comprehension}
\author{Peter D. Turney\\
Institute for Information Technology \\
National Research Council of Canada \\
M-50 Montreal Road \\
Ottawa, Ontario, Canada \\
K1A 0R6 \\
\vspace{1cm}
Phone: (613) 993-8564 \\
Fax: (613) 952-7151 \\
peter.turney@nrc-cnrc.gc.ca}
\date{February 23, 2009}
\begin{document}

\maketitle

\begin{center}
{\Large Analogy Perception Applied to Seven Tests of \\
Word Comprehension}
\end{center}

\vspace{0.5cm}

\begin{abstract}
It has been argued that analogy is the core of cognition. In AI research,
algorithms for analogy are often limited by the need for hand-coded high-level
representations as input. An alternative approach is to use high-level
perception, in which high-level representations are automatically generated
from raw data. Analogy perception is the process of recognizing analogies using
high-level perception. We present PairClass, an algorithm for analogy perception
that recognizes lexical proportional analogies using representations that are
automatically generated from a large corpus of raw textual data. A proportional
analogy is an analogy of the form $A$:$B$::$C$:$D$, meaning ``$A$ is to $B$ as
$C$ is to $D$''. A lexical proportional analogy is a proportional analogy with words,
such as carpenter:wood::mason:stone. PairClass represents the semantic relations
between two words using a high-dimensional feature vector, in which the elements
are based on frequencies of patterns in the corpus. PairClass recognizes analogies
by applying standard supervised machine learning techniques to the feature vectors.
We show how seven different tests of word comprehension can be framed as problems
of analogy perception and we then apply PairClass to the seven resulting sets of
analogy perception problems. We achieve competitive results on all seven tests.
This is the first time a uniform approach has handled such a range of tests of word
comprehension.
\end{abstract}

\vspace{0.5cm}
\noindent \textbf{Keywords:} analogies, word comprehension, test-based AI,
semantic relations, synonyms, antonyms.
\vspace{0.5cm}

\section{Introduction}

Many AI researchers and cognitive scientists believe that analogy is
``the core of cognition'' \cite{hofstadter01}:

\begin{myitemize}

\item ``How do we ever understand anything? Almost always, I think, by using one
or another kind of analogy.'' -- Marvin Minsky \shortcite{minsky86}

\item ``My thesis is this: what makes humans smart is (1) our exceptional ability to
learn by analogy, (2) the possession of symbol systems such as language and
mathematics, and (3) a relation of mutual causation between them whereby
our analogical prowess is multiplied by the possession of relational language.''
-- Dedre Gentner \shortcite{gentner03}

\item ``We have repeatedly seen how analogies and mappings give rise to
secondary meanings that ride on the backs of primary meanings. We have seen
that even primary meanings depend on unspoken mappings, and so in the end,
we have seen that all meaning is mapping-mediated, which is to say, all
meaning comes from analogies.'' -- Douglas Hofstadter \shortcite{hofstadter07}

\end{myitemize}

\noindent These quotes connect analogy with understanding, learning,
language, and meaning. Our research in natural language processing
for word comprehension (lexical semantics) has been guided by
this view of the importance of analogy.

The best-known approach to analogy-making is the Structure-Mapping
Engine (SME) \cite{falkenhainer89}, which is able to process scientific analogies.
SME constructs a mapping between two high-level conceptual representations.
These kinds of high-level analogies are sometimes called \emph{conceptual analogies}.
For example, SME is able to build a mapping between a high-level representation
of Rutherford's model of the atom and a high-level representation of the solar system
\cite{falkenhainer89}. The input to SME consists of hand-coded high-level
representations, written in LISP. (See Appendix~B of Falkenhainer \emph{et al.}
\shortcite{falkenhainer89} for examples of the input LISP code.)

The SME approach to analogy-making has been criticized because it assumes
that hand-coded representations are available as the basic building
blocks for ana\-logy-making \cite{chalmers92}. The process of forming high-level
conceptual representations from raw data (without hand-coding) is called
\emph{high-level perception} \cite{chalmers92}. Turney \shortcite{turney08b}
introduced the Latent Relation Mapping Engine (LRME), which combines ideas from
SME and Latent Relational Analysis (LRA) \cite{turney06}. LRME is
able to construct mappings without hand-coded high-level
representations. Using a kind of high-level perception, LRME builds conceptual
representations from raw data, consisting of a large corpus of plain text,
gathered by a web crawler.

In this paper, we use ideas from LRA and LRME to solve word comprehension
tests. We focus on a kind of lower-level analogy, called
\emph{proportional analogy}, which has the form $A$:$B$::$C$:$D$, meaning
``$A$ is to $B$ as $C$ is to $D$''. Each component mapping in a high-level
conceptual analogy is essentially a lower-level proportional analogy.
For example, in the analogy between the solar system and Rutherford's model
of the atom, the component mappings include the proportional analogies
sun:planet::nucleus:electron and mass:sun::charge:nucleus \cite{turney08b}.

Proportional analogies are common in psychometric tests, such as the
Miller Analogies Test (MAT) and the Graduate Record Examination (GRE).
In these tests, the items in the analogies are usually either geometric figures
or words. An early AI system for proportional analogies with geometric figures
was {\sc Analogy} \cite{evans64} and an early system for words was Argus
\cite{reitman65}. Both of these systems used hand-coded representations to
solve simple proportional analogy questions.

In Section~\ref{sec:analogy-perception}, we present an algorithm we
call PairClass, designed for recognizing proportional analogies with words.
PairClass performs high-level perception \cite{chalmers92}, forming
conceptual representations of semantic relations between words, by
analysis of raw textual data, without hand-coding. The representations
are high-dimensional vectors, in which the values of the elements
are derived from the frequencies of patterns in textual data.
This form of representation is similar to latent semantic analysis (LSA)
\cite{landauer97}, but vectors in LSA represent the meaning of individual
words, whereas vectors in PairClass represent the relations between two words.
The use of frequency vectors to represent semantic relations was introduced
in Turney \emph{et al.} \shortcite{turneyetal03}.

PairClass uses a standard supervised machine learning algorithm
\cite{platt98,witten99} to classify word pairs according to their semantic
relations. A proportional analogy such as sun:planet::nucleus:electron
asserts that the semantic relations between sun and planet are
similar to the semantic relations between nucleus and electron.
The planet orbits the sun; the electron orbits the nucleus. The sun's
gravity attracts the planet; the nucleus's charge attracts the
electron. The task of perceiving this proportional analogy can
be framed as the task of learning to classify sun:planet and
nucleus:electron into the same class, which we might call orbited:orbiter.
Thus our approach to analogy perception is to frame it as a problem of
classification of word pairs (hence the name PairClass).

To evaluate PairClass, we use seven word comprehension tests. This could
be seen as a return to the 1960's psychometric test-based
approach of {\sc Analogy} \cite{evans64}
and Argus \cite{reitman65}, but the difference is that PairClass
achieves human-level scores on the tests without using hand-coded
representations. We believe that word comprehension tests serve as an excellent
benchmark for evaluating progress in computational linguistics.
More generally, we support test-based AI research \cite{bringsjord03}.

In Section~\ref{sec:experiments}, we present our experiments with seven tests:

\begin{myitemize}

\item 374 multiple-choice analogy questions from the SAT college entrance test
\cite{turneyetal03},

\item 80 multiple-choice synonym questions from the TOEFL
(test of English as a foreign language) \cite{landauer97},

\item 50 multiple-choice synonym questions from an ESL (English as a second
language) practice test \cite{turney01},

\item 136 synonym-antonym questions collected from several ESL practice tests
(introduced here),

\item 160 synonym-antonym questions from research in computational linguistics
\cite{lin03},

\item 144 similar-associated-both questions that were used for research in
cognitive psychology \cite{chiarello90}, and

\item 600 noun-modifier relation classification problems from research in
computational linguistics \cite{nastase03}.

\end{myitemize}

We discuss the results of the experiments in Section~\ref{sec:discussion}.
For five of the seven tests, there are past results
that we can compare with the performance of PairClass. In general,
PairClass is competitive, but not the best system. However, the
strength of PairClass is that it is able to handle seven
different tests. As far as we know, no other system can handle this
range of tests. PairClass performs well, although it is competing against
specialized algorithms, developed for single tasks. We believe that
this illustrates the power of analogy perception as a unified approach
to lexical semantics.

Related work is examined in Section~\ref{sec:related}. PairClass is
similar to past work on semantic relation classification
\cite{rosario01,nastase03,turneylittman05,girju07}. For example,
with noun-modifier classification, the task is to classify a
noun-modifier pair, such as \emph{laser printer}, according to the
semantic relation between the head noun, \emph{printer}, and the
modifier, \emph{laser}. In this case, the relation is \emph{instrument:agency}:
the laser is an instrument that is used by the printer. The standard
approach to semantic relation classification is to use supervised machine
learning techniques to classify feature vectors that represent relations.
We demonstrate in this paper that the paradigm of semantic relation
classification can be extended beyond the usual relations, such as
\emph{instrument:agency}, to include analogy, synonymy, antonymy, similarity,
and association.

Limitations and future work are considered in Section~\ref{sec:limitations}.
Limitations of PairClass are the need for a large corpus and the time
required to run the algorithm. We conclude in Section~\ref{sec:conclusion}.

PairClass was briefly introduced in Turney \shortcite{turney08a}. The current
paper describes PairClass in more detail, provides more background information
and discussion, and brings the number of tests up from four to seven.

\section{Analogy Perception}
\label{sec:analogy-perception}

A lexical analogy, $A$:$B$::$C$:$D$, asserts that $A$ is to $B$
as $C$ is to $D$; for example, carpenter:wood::mason:stone
asserts that carpenter is to wood as mason is to stone;
that is, the semantic relations between carpenter and wood are
highly similar to the semantic relations between mason and
stone. In this paper, we frame the task of recognizing lexical analogies
as a problem of classifying word pairs (see Table~\ref{tab:pair-classes}).

\begin{table}[htb]
\centering
\begin{tabular*}{0.55\linewidth}{@{\extracolsep{\fill}}ll}
\hline
Word pair & Class label \\
\hline
carpenter:wood     & artisan:material \\
mason:stone        & artisan:material \\
potter:clay        & artisan:material \\
glassblower:glass  & artisan:material \\
sun:planet         & orbited:orbiter \\
nucleus:electron   & orbited:orbiter \\
earth:moon         & orbited:orbiter \\
starlet:paparazzo  & orbited:orbiter \\
\hline
\end{tabular*}
\caption {Examples of how the task of recognizing lexical analogies may be
viewed as a problem of classifying word pairs.}
\label{tab:pair-classes}
\end{table}

We approach this task as a standard classification problem for supervised
machine learning \cite{witten99}. PairClass takes as input a training set of word
pairs with class labels and a testing set of word pairs without labels.
Each word pair is represented as a vector in a feature space and a supervised
learning algorithm is used to classify the feature vectors. The elements in
the feature vectors are based on the frequencies of automatically defined
patterns in a large corpus. The output of the algorithm is an assignment
of labels to the word pairs in the testing set. For some of the following
experiments, we select a unique label for each word pair; for other
experiments, we assign probabilities to each possible label for each word pair.

For a given word pair, such as mason:stone, the first step is to
generate morphological variations, such as masons:stones. In the
following experiments, we use \emph{morpha} (morphological analyzer) and
\emph{morphg} (morphological generator) for morphological processing
\cite{minnen01}.\footnote{http://www.informatics.susx.ac.uk/research/groups/nlp/carroll/morph.html.}

The second step is to search in a large corpus for phrases of
the following forms:

\begin{myitemize}
\item ``[0 to 1 words] $X$ [0 to 3 words] $Y$ [0 to 1 words]''
\item ``[0 to 1 words] $Y$ [0 to 3 words] $X$ [0 to 1 words]''
\end{myitemize}

\noindent In these templates, $X$:$Y$ consists of morphological variations
of the given word pair; for example, mason:stone, mason:stones, masons:stones,
and so on. Typical phrases for mason:stone would be ``the mason cut the stone with''
and ``the stones that the mason used''. We then normalize all of the
phrases that are found, by using \emph{morpha} to remove suffixes.

The templates we use here are similar to those in Turney \shortcite{turney06},
but we have added extra context words before the first variable ($X$ in the
first template and $Y$ in the second) and after the second variable.
Our morphological processing also differs from Turney \shortcite{turney06}.
In the following experiments, we search in a corpus of $5 \times 10^{10}$
words (about 280 GB of plain text), consisting of web pages gathered
by a web crawler.\footnote{The corpus was collected by Charles Clarke at
the University of Waterloo. We can provide copies of the corpus on request.}
To retrieve phrases from the corpus, we use Wumpus \cite{buettcher05}, an
efficient search engine for passage retrieval
from large corpora.\footnote{http://www.wumpus-search.org/.}

The next step is to generate patterns from all of the phrases that
were found for all of the input word pairs (from both the training
and testing sets). To generate patterns from a phrase, we replace the given
word pairs with variables, $X$ and $Y$, and we replace the remaining words
with a wild card symbol (an asterisk) or leave them as they are.
For example, the phrase ``the mason cut the stone with'' yields the
patterns ``the $X$ cut * $Y$ with'', ``* $X$ * the $Y$ *'', and so on.
If a phrase contains $n$ words, then it yields $2^{(n-2)}$ patterns.

Each pattern corresponds to a feature in the feature vectors that we
will generate. Since a typical input set of word pairs yields
millions of patterns, we need to use feature selection, to reduce
the number of patterns to a manageable quantity. For each pattern,
we count the number of input word pairs that generated the pattern.
For example, ``* $X$ cut * $Y$ *'' is generated by both mason:stone
and carpenter:wood. We then sort the patterns in descending order
of the number of word pairs that generated them. If there are $N$
input word pairs (and thus $N$ feature vectors, including both the
training and testing sets), then we select
the top $kN$ patterns and drop the remainder. In the
following experiments, $k$ is set to 20. The algorithm is not
sensitive to the precise value of $k$.

The reasoning behind the feature selection algorithm is that shared
patterns make more useful features than rare patterns.
The number of features ($kN$) depends on the number of word pairs ($N$),
because, if we have more feature vectors, then we need more features to
distinguish them. Turney \shortcite{turney06} also selects patterns
based on the number of pairs that generate them, but the number of
selected patterns is a constant (8000), independent of the number of
input word pairs.

The next step is to generate feature vectors, one vector for each
input word pair. Each of the $N$ feature vectors has $kN$ elements,
one element for each selected pattern. The value of an element
in a vector is given by the logarithm of the frequency in the corpus
of the corresponding pattern for the given word pair. For example,
suppose the given pair is mason:stone and the pattern is
``* $X$ cut * $Y$ *''. We look at the normalized phrases that
we collected for mason:stone and we count how many match this
pattern. If $f$ phrases match the pattern, then the value of
this element in the feature vector is $\log(f+1)$ (we add $1$
because $\log(0)$ is undefined). Each feature vector is then
normalized to unit length. The normalization ensures that
features in vectors for high-frequency word pairs are
comparable to features in vectors for low-frequency word pairs.

Table~\ref{tab:features} shows the first and last ten features (excluding
zero-valued features) and the corresponding feature values
for the word pair audacious:boldness, taken from the SAT analogy
questions. The features are in descending order of the number
of word pairs that generate them; that is, they are ordered from
common to rare. Thus the first features typically involve patterns
with many wild cards and high-frequency words, and the first
feature values are usually nonzero. The last features often have
few wild cards and contain low-frequency words, with feature
values that are usually zero. The feature vectors are generally
highly sparse (i.e., they are mainly zeros; if $f = 0$, then
$\log(f+1) = 0$).

\begin{table}[htb]
\centering
\begin{tabular}{rlc}
\hline
Feature number  &  Feature (pattern)                  &  Value (normalized log) \\
\hline
1               &  ``* $X$ * * $Y$ *''                &  0.090 \\
2               &  ``* $Y$ * * $X$ *''                &  0.150 \\
3               &  ``* $X$ * $Y$ *''                  &  0.198 \\
4               &  ``* $Y$ * $X$ *''                  &  0.221 \\
5               &  ``* $X$ * * * $Y$ *''              &  0.045 \\
7               &  ``* $X$ $Y$ *''                    &  0.233 \\
8               &  ``* $Y$ $X$ *''                    &  0.167 \\
10              &  ``* $Y$ * the $X$ *''              &  0.071 \\
12              &  ``* $Y$ and * $X$ *''              &  0.116 \\
13              &  ``* $X$ and $Y$ *''                &  0.135 \\
\hline
27,591          &  ``define $X$ * $Y$ *''             &  0.045 \\
28,524          &  ``what $Y$ and $X$ *''             &  0.045 \\
28,804          &  ``for $Y$ and * $X$ and''          &  0.045 \\
29,017          &  ``very $X$ and $Y$ *''             &  0.045 \\
32,028          &  ``s $Y$ and $X$ and''              &  0.045 \\
34,893          &  ``understand $X$ * $Y$ *''         &  0.071 \\
35,027          &  ``* $X$ be not * $Y$ but''         &  0.045 \\
39,410          &  ``* $Y$ and $X$ cause''            &  0.045 \\
41,303          &  ``* $X$ but $Y$ and''              &  0.105 \\
43,511          &  ``be $X$ not $Y$ *''               &  0.105 \\
\hline
\end{tabular}
\caption {The first and last ten features, excluding zero-valued
features, for the pair $X$:$Y$ $=$ audacious:boldness. (The ``s''
in the pattern for feature 32,028 is part of a possessive noun.
The ``be'' in the patterns for features 35,027 and 43,511
is the result of normalizing ``is'' and ``was'' with \emph{morpha}.)}
\label{tab:features}
\end{table}

Now that we have a feature vector for each input word pair,
we can apply a standard supervised learning algorithm.
In the following experiments, we use a sequential minimal optimization
(SMO) support vector machine (SVM) with a radial basis function (RBF)
kernel \cite{platt98}, as implemented in Weka (Waikato Environment for
Knowledge Analysis) \cite{witten99}.\footnote{http://www.cs.waikato.ac.nz/ml/weka/.}
The algorithm generates probability estimates for each class by fitting logistic
regression models to the outputs of the SVM. We disable the normalization option
in Weka, since the vectors are already normalized to unit length.
We chose the SMO RBF algorithm because it is fast, robust, and
it easily handles large numbers of features.

In the following experiments, PairClass is applied to each of the seven
tests with no adjustments or tuning of the learning parameters to the specific
problems. Some work is required to fit each problem into the general framework
of PairClass (analogy perception: supervised classification of word pairs), but
the core algorithm is the same in each case.

It might be objected that what PairClass does should not be considered
as high-level perception, in the sense given by Chalmers \emph{et al.}.
\shortcite{chalmers92}. They define high-level perception as follows:

\begin{quote}
Perceptual processes form a spectrum, which for convenience we can divide
into two components. ... [We] have low-level perception, which involves the
early processing of information from the various sensory modalities. High-level
perception, on the other hand, involves taking a more global view of this
information, extracting \emph{meaning} from the raw material by accessing concepts,
and making sense of situations at a conceptual level. This ranges from the recognition
of objects to the grasping of abstract relations, and on to understanding
entire situations as coherent wholes. ... The study of high-level perception
leads us directly to the problem of mental \emph{representation}. Representations
are the fruits of perception.
\end{quote}

Spoken or written language can be converted to electronic text
by speech recognition software or optical character recognition software.
It seems reasonable to call this low-level perception. PairClass takes
electronic text as input and generates high-dimensional feature vectors
from the text. These feature vectors represent abstract semantic relations
and they can be used to classify semantic relations into various semantic
classes. It seems reasonable to call this high-level perception.
We do not claim that PairClass has the richness and complexity of human
high-level perception, but it is nonetheless a (simple, restricted) form of
high-level perception.

\section{Experiments}
\label{sec:experiments}

This section presents seven sets of experiments. We explain how each
of the seven tests is treated as a problem of analogy perception,
we give the experimental results, and we discuss past work
with each test.

\subsection{SAT Analogies}
\label{subsec:sat}

In this section, we apply PairClass to the task of recognizing lexical
analogies. To evaluate the performance, we use a set of 374 multiple-choice
questions from the SAT college entrance exam. Table~\ref{tab:sat} shows
a typical question. The target pair is called the \emph{stem}.
The task is to select the choice pair that is most analogous to
the stem pair.

\begin{table}[htb]
\centering
\begin{tabular*}{0.5\linewidth}{@{\extracolsep{\fill}}lll}
\hline
Stem:      &       & mason:stone \\
\hline
Choices:   & (a)   & teacher:chalk \\
           & (b)   & carpenter:wood \\
           & (c)   & soldier:gun \\
           & (d)   & photograph:camera \\
           & (e)   & book:word \\
\hline
 Solution: & (b)   & carpenter:wood \\
\hline
\end{tabular*}
\caption {An example of a question from the 374 SAT analogy questions.}
\label{tab:sat}
\end{table}

The problem of recognizing
lexical analogies was first attempted with a system called Argus \cite{reitman65},
using a small hand-built semantic network with a spreading activation
algorithm. Turney \emph{et al.} \shortcite{turneyetal03} used a combination of 13
independent modules. Veale \shortcite{veale04} used a spreading activation
algorithm with WordNet (in effect, treating WordNet as a semantic network).
Turney \shortcite{turney05} used a corpus-based algorithm.

We may view Table~\ref{tab:sat} as a binary classification problem, in which
mason:stone and carpenter:wood are positive examples and the remaining word
pairs are negative examples. The difficulty is that the labels of the
choice pairs must be hidden from the learning algorithm. That is, the
training set consists of one positive example (the stem pair) and the
testing set consists of five unlabeled examples (the five choice pairs).
To make this task more tractable, we randomly choose a stem pair from
one of the 373 other SAT analogy questions, and we assume that this
new stem pair is a negative example, as shown in Table~\ref{tab:sat-frame}.

\begin{table}[htb]
\centering
\begin{tabular*}{0.6\linewidth}{@{\extracolsep{\fill}}lll}
\hline
Word pair & Train or test & Class label \\
\hline
mason:stone        & train               & positive \\
tutor:pupil        & train               & negative \\
\hline
teacher:chalk      & test                & hidden \\
carpenter:wood     & test                & hidden \\
soldier:gun        & test                & hidden \\
photograph:camera  & test                & hidden \\
book:word          & test                & hidden \\
\hline
\end{tabular*}
\caption {How to fit a SAT analogy question into the framework of
supervised classification of word pairs. The randomly chosen stem
pair is tutor:pupil.}
\label{tab:sat-frame}
\end{table}

To answer a SAT question, we use PairClass to estimate
the probability that each testing example is positive, and we guess
the testing example with the highest probability.
Learning from a training set with only one positive example and one
negative example is difficult, since the learned model can be highly unstable.
To increase the stability, we repeat the learning process 10 times,
using a different randomly chosen negative training example each time.
For each testing word pair, the 10 probability estimates are averaged together.
This is a form of bagging \cite{breiman96}. Table~\ref{tab:sat-solved}
shows an example of an analogy that has been correctly solved by PairClass.

\begin{table}[htb]
\centering
\begin{tabular*}{0.85\linewidth}{@{\extracolsep{\fill}}llll}
\hline
Stem:      &       & insubordination:punishment & Probability\\
\hline
Choices:   & (a)   & evening:night              & 0.236 \\
           & (b)   & earthquake:tornado         & 0.260 \\
           & (c)   & candor:falsehood           & 0.391 \\
           & (d)   & heroism:praise             & 0.757 \\
           & (e)   & fine:penalty               & 0.265 \\
\hline
 Solution: & (d)   & heroism:praise             & 0.757 \\
\hline
\end{tabular*}
\caption {An example of a correctly solved SAT analogy question.}
\label{tab:sat-solved}
\end{table}

PairClass attains an accuracy of 52.1\% on the 374 SAT analogy questions.
The best previous result is an accuracy of 56.1\% \cite{turney05}.
Random guessing would yield an accuracy of 20\% (five choices per question).
The average senior high school student achieves 57\% correct
\cite{turney06}. The \emph{ACL Wiki} lists 12 previously published results with the
374 SAT analogy questions.\footnote{For more information, see \emph{SAT Analogy
Questions (State of the art)} at http://aclweb.org/aclwiki/. There were 12
previous results at the time of writing, but the list is likely to grow.}
Adding PairClass to the list, we have 13 results. PairClass has
the third highest accuracy of the 13 systems.

\subsection{TOEFL Synonyms}

Now we apply PairClass to the task of recognizing synonyms,
using a set of 80 multiple-choice synonym questions from the
TOEFL (test of English as a foreign language). A sample question
is shown in Table~\ref{tab:toefl}. The task is to select the choice
word that is most similar in meaning to the stem word.

\begin{table}[htb]
\centering
\begin{tabular*}{0.4\linewidth}{@{\extracolsep{\fill}}lll}
\hline
Stem:      &       & levied \\
\hline
Choices:   & (a)   & imposed \\
           & (b)   & believed \\
           & (c)   & requested \\
           & (d)   & correlated \\
\hline
 Solution: & (a)   & imposed \\
\hline
\end{tabular*}
\caption {An example of a question from the 80 TOEFL synonym questions.}
\label{tab:toefl}
\end{table}

Synonymy can be viewed
as a high degree of semantic similarity. The most common way to measure
semantic similarity is to measure the distance between words in WordNet
\cite{resnik95,jiang97,hirst98,budanitsky01}. Corpus-based measures
of word similarity are also common \cite{lesk69,landauer97,turney01}.

We may view Table~\ref{tab:toefl} as a binary classification problem,
in which the pair levied:imposed is a positive example of the class
\emph{synonymous} and the other possible pairings are negative
examples, as shown in Table~\ref{tab:toefl-frame}.

\begin{table}[htb]
\centering
\begin{tabular*}{0.45\linewidth}{@{\extracolsep{\fill}}lll}
\hline
Word pair & Class label \\
\hline
levied:imposed      & positive \\
levied:believed     & negative \\
levied:requested    & negative \\
levied:correlated   & negative \\
\hline
\end{tabular*}
\caption {How to fit a TOEFL synonym question into the framework of
supervised classification of word pairs.}
\label{tab:toefl-frame}
\end{table}

The 80 TOEFL questions yield 320 ($80 \times 4$) word pairs,
80 labeled positive and 240 labeled negative. We apply PairClass
to the word pairs using ten-fold cross-validation. In each
random fold, 90\% of the pairs are used for training and 10\% are
used for testing. For each fold, we use the learned model to assign
probabilities to the testing pairs. Our guess for each TOEFL question
is the choice that has the highest probability of being positive, when
paired with the corresponding stem. Table~\ref{tab:toefl-solved} gives an
example of a correctly solved question.

\begin{table}[htb]
\centering
\begin{tabular*}{0.65\linewidth}{@{\extracolsep{\fill}}llll}
\hline
Stem:      &       & prominent     & Probability \\
\hline
Choices:   & (a)   & battered      & 0.005 \\
           & (b)   & ancient       & 0.114 \\
           & (c)   & mysterious    & 0.010 \\
           & (d)   & conspicuous   & 0.998 \\
\hline
 Solution: & (d)   & conspicuous   & 0.998 \\
\hline
\end{tabular*}
\caption {An example of a correctly solved TOEFL synonym question.}
\label{tab:toefl-solved}
\end{table}

PairClass attains an accuracy of 76.2\%. For comparison, the
\emph{ACL Wiki} lists 15 previously published results with the 80 TOEFL
synonym questions.\footnote{See \emph{TOEFL Synonym Questions (State of the
art)} at http://aclweb.org/aclwiki/. There were 15 systems at the time of
writing, but the list is likely to grow.} Adding PairClass to the list,
we have 16 algorithms. PairClass has the ninth highest accuracy of
the 16 systems. The best previous result is an accuracy of
97.5\% \cite{turneyetal03}, obtained using a hybrid of four different
algorithms. Random guessing would yield an accuracy of 25\% (four choices
per question). The average foreign applicant to a US university achieves
64.5\% correct \cite{landauer97}.

\subsection{ESL Synonyms}

The 50 ESL synonym questions are similar to the TOEFL synonym questions,
except that each question includes a sentence that shows the stem word in
context. Table~\ref{tab:esl} gives an example. In our experiments, we ignore
the sentence context and treat the ESL synonym questions the same way as
we treated the TOEFL synonym questions (see Table~\ref{tab:esl-frame}).

\begin{table}[htb]
\centering
\begin{tabular*}{0.6\linewidth}{@{\extracolsep{\fill}}lll}
\hline
Stem:     &       &   \parbox[t]{1.7in}{\raggedright ``A \emph{rusty}
nail is not as strong as a clean, new one.''} \\
\hline
Choices:  & (a)   &   corroded \\
          & (b)   &   black \\
          & (c)   &   dirty \\
          & (d)   &   painted \\
\hline
Solution: & (a)   & corroded \\
\hline
\end{tabular*}
\caption {An example of a question from the 50 ESL synonym questions.}
\label{tab:esl}
\end{table}

\begin{table}[htb]
\centering
\begin{tabular*}{0.45\linewidth}{@{\extracolsep{\fill}}lll}
\hline
Word pair & Class label \\
\hline
rusty:corroded      & positive \\
rusty:black         & negative \\
rusty:dirty         & negative \\
rusty:painted       & negative \\
\hline
\end{tabular*}
\caption {How to fit an ESL synonym question into the framework of
supervised classification of word pairs.}
\label{tab:esl-frame}
\end{table}

The 50 ESL questions yield 200 ($50 \times 4$) word pairs, 50 labeled
positive and 150 labeled negative. We apply PairClass to the word
pairs using ten-fold cross-validation. Our guess for each question
is the choice word that has the highest probability of being positive, when
paired with the corresponding stem word.

PairClass attains an accuracy of 78.0\%. The best previous result
is 82.0\% \cite{jarmasz03}. The \emph{ACL Wiki} lists 8 previously
published results for the 50 ESL synonym questions.\footnote{See
\emph{ESL Synonym Questions (State of the art)} at http://aclweb.org/aclwiki/.
There were 8 systems at the time of writing, but the list is likely to grow.}
Adding PairClass to the list, we have 9 algorithms. PairClass has the third
highest accuracy of the 9 systems. The average human score is unknown.
Random guessing would yield an accuracy of 25\% (four choices
per question).

\subsection{ESL Synonyms and Antonyms}

The task of classifying word pairs as either synonyms or antonyms
readily fits into the framework of supervised classification of word pairs.
Table~\ref{tab:esl-antonyms} shows some examples from a set of
136 ESL (English as a second language) practice questions that
we collected from various ESL websites.

\begin{table}[htb]
\centering
\begin{tabular*}{0.5\linewidth}{@{\extracolsep{\fill}}ll}
\hline
Word pair & Class label \\
\hline
galling:irksome             &  synonyms \\
yield:bend                  &  synonyms \\
naive:callow                &  synonyms \\
advise:suggest              &  synonyms \\
dissimilarity:resemblance   &  antonyms \\
commend:denounce            &  antonyms \\
expose:camouflage           &  antonyms \\
unveil:veil                 &  antonyms \\
\hline
\end{tabular*}
\caption {Examples of synonyms and antonyms from 136 ESL practice questions.}
\label{tab:esl-antonyms}
\end{table}

Hatzivassiloglou and McKeown \shortcite{hatzivassiloglou97}
propose that antonyms and synonyms can be distinguished by their
semantic orientation. A word that suggests praise has a positive
semantic orientation, whereas criticism is negative semantic orientation.
Antonyms tend to have opposite semantic orientation (fast:slow is
positive:negative) and synonyms tend to have the same semantic orientation
(fast:quick is positive:positive). However, this proposal has not been evaluated,
and it is not difficult to find counter-examples (simple:simplistic
is positive:negative, yet the words are synonyms, rather than antonyms).

Lin \emph{et al.} \shortcite{lin03} distinguish synonyms from antonyms using
two patterns, ``from $X$ to $Y$'' and ``either $X$ or $Y$''. When $X$ and $Y$
are antonyms, they occasionally appear in a large corpus in one of these
two patterns, but it is very rare for synonyms to appear in these patterns.
Our approach is similar to Lin \emph{et al.} \shortcite{lin03}, but we do not
rely on hand-coded patterns; instead, PairClass patterns are generated
automatically.

Using ten-fold cross-validation, PairClass attains an accuracy of
75.0\%. Always guessing the majority class would result in an accuracy
of 65.4\%. The average human score is unknown and there are no
previous results for comparison.

\subsection{CL Synonyms and Antonyms}

To compare PairClass with the algorithm of Lin \emph{et al.} \shortcite{lin03},
this experiment uses their set of 160 word pairs, 80 labeled
\emph{synonym} and 80 labeled \emph{antonym}. These 160 pairs
were chosen by Lin \emph{et al.} \shortcite{lin03} for their high frequency;
thus they are somewhat easier to classify than the 136 ESL practice questions.
Some examples are given in Table~\ref{tab:cl-antonyms}.

\begin{table}[htb]
\centering
\begin{tabular*}{0.6\linewidth}{@{\extracolsep{\fill}}ll}
\hline
Word pair & Class label \\
\hline
audit:review                  &  synonyms \\
education:tuition             &  synonyms \\
location:position             &  synonyms \\
material:stuff                &  synonyms \\
ability:inability             &  antonyms \\
balance:imbalance             &  antonyms \\
exaggeration:understatement   &  antonyms \\
inferiority:superiority       &  antonyms \\
\hline
\end{tabular*}
\caption {Examples of synonyms and antonyms from 160 labeled pairs
for experiments in computational linguistics (CL).}
\label{tab:cl-antonyms}
\end{table}

Lin \emph{et al.} \shortcite{lin03} report their performance using
precision (86.4\%) and recall (95.0\%), instead of accuracy,
but an accuracy of 90.0\% can be derived from their figures,
with some minor algebraic manipulation. Using ten-fold cross-validation,
PairClass has an accuracy of 81.9\%. Random guessing would
yield an accuracy of 50\%. The average human score is unknown.

\subsection{Similar, Associated, and Both}

A common criticism of corpus-based measures of word similarity (as
opposed to lexicon-based measures) is that they are merely detecting
associations (co-oc\-cur\-rences), rather than actual semantic similarity
\cite{lund95}. To address this criticism, Lund \emph{et al.} \shortcite{lund95}
evaluated their algorithm for measuring word similarity with
word pairs that were labeled \emph{similar}, \emph{associated},
or \emph{both}. These labeled pairs were originally created for
cognitive psychology experiments with human subjects \cite{chiarello90}.
Table~\ref{tab:associated} shows some examples from this
collection of 144 word pairs (48 pairs in each of the three classes).

\begin{table}[htb]
\centering
\begin{tabular*}{0.4\linewidth}{@{\extracolsep{\fill}}ll}
\hline
Word pair & Class label \\
\hline
table:bed      & similar \\
music:art      & similar \\
hair:fur       & similar \\
house:cabin    & similar \\
cradle:baby    & associated \\
mug:beer       & associated \\
camel:hump     & associated \\
cheese:mouse   & associated \\
ale:beer       & both \\
uncle:aunt     & both \\
pepper:salt    & both \\
frown:smile    & both \\
\hline
\end{tabular*}
\caption {Examples of word pairs labeled \emph{similar}, \emph{associated},
or \emph{both}.}
\label{tab:associated}
\end{table}

Lund \emph{et al.} \shortcite{lund95} did not measure the accuracy of their
algorithm on this three-class classification problem. Instead, following
standard practice in cognitive psychology, they showed that their
algorithm's similarity scores for the 144 word pairs were correlated
with the response times of human subjects in priming tests. In a typical
priming test, a human subject reads a \emph{priming} word (\emph{cradle}) and
is then asked to complete a partial word (complete \emph{bab} as \emph{baby})
or to distinguish a word (\emph{baby}) from a non-word (\emph{baol}).
The time required to perform the task is taken to indicate the strength
of the cognitive link between the two words (\emph{cradle} and \emph{baby}).

Using ten-fold cross-validation, PairClass attains an accuracy of
77.1\% on the 144 word pairs. Since the three classes are of equal size,
guessing the majority class and random guessing both yield an
accuracy of 33.3\%. The average human score is unknown and there are no
previous results for comparison.

\subsection{Noun-Modifier Relations}

A noun-modifier expression is a compound of two (or more) words, a
head noun and a modifier of the head. The modifier is usually a noun
or adjective. For example, in the noun-modifier expression
\emph{student discount}, the head noun \emph{discount} is modified by the
noun \emph{student}.

Noun-modifier expressions are very common in English. There is
wide variation in the types of semantic relations between heads and
modifiers. A challenging task for natural language processing
is to classify noun-modifier pairs according to their semantic relations.
For example, in the noun-modifier expression \emph{electron microscope},
the relation might be \emph{theme:tool} (a microscope for electrons; perhaps for
viewing electrons), \emph{instrument:agency} (a microscope that uses electrons), or
\emph{material:artifact} (a microscope made out of electrons).\footnote{The
correct answer is \emph{instrument:agency}.} There are many potential
applications for algorithms that can automatically classify noun-modifier
pairs according to their semantic relations.

Nastase and Szpakowicz \shortcite{nastase03} collected 600 noun-modifier
pairs and hand-labeled them with 30 different classes of semantic relations.
The 30 classes were organized into five groups: causality, temporality,
spatial, participant, and quality. Due to the difficulty of distinguishing
30 classes, most researchers prefer to treat this as a five-class classification
problem. Table~\ref{tab:noun-modifiers} shows some examples of
noun-modifier pairs with the five-class labels.

\begin{table}[htb]
\centering
\begin{tabular*}{0.5\linewidth}{@{\extracolsep{\fill}}ll}
\hline
Word pair              &   Class label \\
\hline
cold:virus             &   causality \\
onion:tear             &   causality \\
morning:frost          &   temporality \\
late:supper            &   temporality \\
aquatic:mammal         &   spatial \\
west:coast             &   spatial \\
dream:analysis         &   participant \\
police:intervention    &   participant \\
copper:coin            &   quality \\
rice:paper             &   quality \\
\hline
\end{tabular*}
\caption {Examples of noun-modifier word pairs labeled with five semantic relations.}
\label{tab:noun-modifiers}
\end{table}

The design of the PairClass algorithm is closely related to past work on the
problem of classifying noun-modifier semantic relations, so we will examine
this past work in more detail than in our discussions of related work for the other
six tests. Section~\ref{sec:related} will focus on the relation between PairClass
and past work on semantic relation classification.

Using ten-fold cross-validation, PairClass achieves an accuracy of
58.0\% on the task of classifying the 600 noun-modifier pairs into
five classes. The best previous result was also 58.0\% \cite{turney06}.
The \emph{ACL Wiki} lists 5 previously published results with the
600 noun-modifier pairs.\footnote{See \emph{Noun-Modifier Semantic Relations
(State of the art)} at http://aclweb.org/aclwiki/. There were 5 systems
at the time of writing, but the list is likely to grow.} Adding PairClass
to the list, we have 6 algorithms. PairClass ties for first place in the
set of 6 systems. Guessing the majority class would result in an accuracy
of 43.3\%. The average human score is unknown.

\section{Discussion}
\label{sec:discussion}

The seven experiments are summarized in Tables \ref{tab:summary-tasks}
and \ref{tab:summary-results}. For the five experiments for which there are previous
results, PairClass is not the best, but it performs competitively. For the other
two experiments, PairClass performs significantly above the baselines.
However, the strength of this approach is not its performance on any one task,
but the range of tasks it can handle. No other algorithm has been applied to
this range of lexical semantic problems.

\begin{table}[htb]
\centering
\begin{tabular*}{\textwidth}{@{\extracolsep{\fill}}lrrr}
\hline
Experiment                      & Vectors      & Features         & Classes \\
\hline
SAT Analogies                   & 2,244        & 44,880           & 374 \\
TOEFL Synonyms                  & 320          & 6,400            & 2 \\
ESL Synonyms                    & 200          & 4,000            & 2 \\
ESL Synonyms and Antonyms       & 136          & 2,720            & 2 \\
CL Synonyms and Antonyms        & 160          & 3,200            & 2 \\
Similar, Associated, and Both   & 144          & 2,880            & 3 \\
Noun-Modifier Relations         & 600          & 12,000           & 5 \\
\hline
\end{tabular*}
\caption {Summary of the seven tasks. See Section~\ref{sec:experiments} for
explanations. The number of features is 20 times the number of vectors, as mentioned
in Section~\ref{sec:analogy-perception}. For SAT Analogies, the number of vectors
is $374 \times 6$. For TOEFL Synonyms, the number of vectors is $80 \times 4$.
For ESL Synonyms, the number of vectors is $50 \times 4$.}
\label{tab:summary-tasks}
\end{table}

\begin{table}[htb]
\centering
\begin{tabular*}{\textwidth}{@{\extracolsep{\fill}}lcccc}
\hline
Experiment                      & Accuracy & Best previous & Baseline   & Rank \\
\hline
SAT Analogies                   & 52.1\%   & 56.1\%        &  20.0\%    & 3 of 13 \\
TOEFL Synonyms                  & 76.2\%   & 97.5\%        &  25.0\%    & 9 of 16 \\
ESL Synonyms                    & 78.0\%   & 82.0\%        &  25.0\%    & 3 of 9 \\
ESL Synonyms and Antonyms       & 75.0\%   & -             &  65.4\%    & - \\
CL Synonyms and Antonyms        & 81.9\%   & 90.0\%        &  50.0\%    & 2 of 2 \\
Similar, Associated, and Both   & 77.1\%   & -             &  33.3\%    & - \\
Noun-Modifier Relations         & 58.0\%   & 58.0\%        &  43.3\%    & 1 of 6 \\
\hline
\end{tabular*}
\caption {Summary of experimental results. See Section~\ref{sec:experiments}
for explanations. For the Noun-Modifier Relations, PairClass is tied for first place.}
\label{tab:summary-results}
\end{table}

Of the seven tests we use here, as far as we know, only the noun-modifier
relations have been approached using a standard supervised learning
algorithm. For the other six tests, PairClass is the first attempt
to apply supervised learning.\footnote{Turney \emph{et al.} \shortcite{turneyetal03}
apply something like supervised learning to the SAT analogies and TOEFL
synonyms, but it would be more accurate to call it reinforcement learning,
rather than standard supervised learning.} The advantage of being able to cast
these six problems in the framework
of standard supervised learning problems is that we can now exploit
the huge literature on supervised learning. Past work on these problems
has required implicitly coding our knowledge of the nature of the
task into the structure of the algorithm. For example, the structure
of the algorithm for latent semantic analysis (LSA) implicitly contains
a theory of synonymy \cite{landauer97}. The problem with this approach
is that it can be very difficult to work out how to modify the algorithm
if it does not behave the way we want. On the other hand, with a supervised
learning algorithm, we can put our knowledge into the labeling of the
feature vectors, instead of putting it directly into the algorithm.
This makes it easier to guide the system to the desired behaviour.

Humans are able to make analogies without supervised learning.
It might be argued that the requirement for supervision is a major
limitation of PairClass. However, with our approach to the SAT analogy
questions (see Section~\ref{subsec:sat}), we are blurring
the line between supervised and unsupervised learning, since the
training set for a given SAT question consists of a single real
positive example (and a single ``virtual'' or ``simulated'' negative example).
In effect, a single example (such as mason:stone in Table~\ref{tab:sat-frame})
becomes a \emph{sui generis}; it constitutes a class of its own. It may be
possible to apply the machinery of supervised learning to other
problems that apparently call for unsupervised learning (for example,
clustering or measuring similarity), by using this \emph{sui generis} device.

\section{Related Work}
\label{sec:related}

One of the first papers using supervised machine learning to classify
word pairs was Rosario and Hearst's \shortcite{rosario01} paper on
classifying noun-modifier pairs in the medical domain.
For example, the noun-modifier expression \emph{brain biopsy} was
classified as \emph{Procedure}. Rosario and Hearst \shortcite{rosario01}
constructed feature vectors for each noun-modifier pair using
MeSH (Medical Subject Headings) and UMLS (Unified Medical Language System)
as lexical resources. They then trained a neural network to distinguish
13 classes of semantic relations, such as \emph{Cause}, \emph{Location},
\emph{Measure}, and \emph{Instrument}. Nastase and Szpakowicz \shortcite{nastase03}
explored a similar approach to classifying general-domain noun-modifier pairs,
using WordNet and Roget's Thesaurus as lexical resources.

Turney and Littman \shortcite{turneylittman05} used corpus-based features
for classifying noun-modifier pairs. Their
features were based on 128 hand-coded patterns. They used a nearest-neighbour
learning algorithm to classify general-domain noun-modifier pairs into
30 different classes of semantic relations. Turney \shortcite{turney05,turney06}
later addressed the same problem using 8000 automatically generated patterns.

One of the tasks in SemEval 2007 was the classification of semantic relations
between nominals \cite{girju07}.\footnote{SemEval 2007 was the Fourth
International Workshop on Semantic Evaluations. More information on Task 4,
the classification of semantic relations between nominals, is available at
http://purl.org/net/semeval/\-task4.}
The problem is to classify semantic relations between nominals (nouns and noun
compounds) in the context of a sentence. The task attracted 14 teams who
created 15 systems, all of which used supervised machine learning with
features that were lexicon-based, corpus-based, or both.

PairClass is most similar to the algorithm of Turney \shortcite{turney06},
but it differs in the following ways:

\begin{myitemize}

\item PairClass does not use a lexicon to find synonyms for the input word
pairs. One of our goals in this paper is to show that a pure corpus-based
algorithm can handle synonyms without a lexicon. This considerably simplifies
the algorithm.

\item PairClass uses a support vector machine (SVM) instead of a nearest
neighbour (NN) learning algorithm.

\item PairClass does not use the singular value decomposition (SVD) to smooth
the feature vectors. It has been our experience that SVD is not necessary with SVMs.

\item PairClass generates probability estimates, whereas Turney \shortcite{turney06}
uses a cosine measure of similarity. Probability estimates can be
readily used in further downstream processing, but cosines are less
useful.

\item The automatically generated patterns in PairClass are slightly
more general than the patterns of Turney \shortcite{turney06}, as
mentioned in Section~\ref{sec:analogy-perception}.

\item The morphological processing in PairClass \cite{minnen01} is more
sophisticated than in Turney \shortcite{turney06}.

\end{myitemize}

\noindent However, we believe that the main contribution of this paper
is not PairClass itself, but the extension of supervised word pair
classification beyond the classification of noun-modifier pairs
and semantic relations between nominals, to analogies, synonyms, antonyms,
and associations. As far as we know, this has not been done before.

\section{Limitations and Future Work}
\label{sec:limitations}

The main limitation of PairClass is the need for a large corpus. Phrases
that contain a pair of words tend to be more rare than phrases that
contain either of the members of the pair, thus
a large corpus is needed to ensure that sufficient numbers of phrases are
found for each input word pair. The size of the corpus has a cost
in terms of disk space and processing time. In the future, as hardware improves,
this will become less of an issue, but there may be ways to improve
the algorithm, so that a smaller corpus is sufficient.

Human language can be creatively extended as needed. Given a newly-defined
word, a human would be able to use it immediately in an analogy.
Since PairClass requires a large number of phrases for each pair of
words, it would be unable to handle a newly-defined word. A problem for
future work is the extension of PairClass, so that it is able to work
with definitions of words. One approach is a hybrid algorithm that
combines a corpus-based algorithm with a lexicon-based algorithm.
For example, Turney \emph{et al.} \shortcite{turneyetal03} describe
an algorithm that combines 13 different modules for solving proportional
analogies with words.

\section{Conclusion}
\label{sec:conclusion}

The PairClass algorithm classifies word pairs according to their
semantic relations, using features generated from a large corpus of text.
We describe PairClass as performing analogy perception, because it
recognizes lexical proportional analogies using a form of high-level
perception \cite{chalmers92}. For given input training and testing sets
of word pairs, it automatically generates patterns and constructs its own
representations of the word pairs as high-dimensional feature vectors.
No hand-coding of representations is involved.

We believe that analogy perception provides a unified approach to
natural language processing for a wide variety of lexical semantic
tasks. We support this by applying PairClass to seven different
tests of word comprehension. It achieves competitive performance on
the tests, although it is competing with algorithms that were developed
for single tasks. More significant is the range of tasks that can
be framed as problems of analogy perception.

The idea of subsuming a broad range of semantic phenomena under analogies
has been suggested by many researchers \cite{minsky86,gentner03,hofstadter07}.
In computational lingistics, analogical algorithms have been applied to
machine translation \cite{lepage05}, morphology \cite{lepage98}, and semantic
relations \cite{turneylittman05}. Analogy provides a framework that
has the potential to unify the field of semantics. This paper is a small
step towards that goal.

In this paper, we have used tests from educational testing (SAT analogies
and TOEFL synonyms), second language practice (ESL synonyms and ESL
synonym and antonyms), computational linguistics (CL synonyms and
antonyms and noun-modifiers), and cognitive psychology (similar, associated,
and both). Six of the tests have been used in previous research and
four of the tests have associated performance results and bibliographies
in the \emph{ACL Wiki}. Shared tests make it possible for researchers to
compare their algorithms and assess the progress of the field.

Applying human tests to machines is a natural way to evaluate progress in AI.
Five of the seven tests were originally developed for humans. For the SAT
and TOEFL tests, the average human scores are available. On the SAT
test, PairClass has an accuracy of 52.1\%, with a 95\% confidence
interval ranging from 46.9\% to 57.3\% (using the Binomial Exact test).
The average senior high school student applying to a US university
achieves 57\% \cite{turney06}, which is within the 95\% confidence
interval for PairClass. On the TOEFL synonym test, PairClass has an
accuracy of 76.2\%, with a 95\% confidence interval ranging from
65.4\% to 85.1\% (using the Binomial Exact test). The average foreign
applicant to a US university achieves 64.5\% \cite{landauer97},
which is below the 95\% confidence interval for PairClass. Thus
PairClass performance on SAT is not significantly different from average
human performance, and PairClass performance on TOEFL is significantly
better than average human performance.

One criticism of AI as a field is that its success stories are
limited to narrow domains, such as chess. Human intelligence
has a generality and flexibility that AI currently lacks. This paper
is a tiny step towards the goal of performing competively on a
wide range of tests, rather than performing very well on a single test.

\section*{Acknowledgements}

Thanks to Michael Littman for the 374 SAT analogy questions, Thomas Landauer
for the 80 TOEFL synonym questions, Donna Tatsuki for the 50 ESL synonym
questions, Dekang Lin for the 160 synonym-antonym questions, Christine Chiarello
for the 144 similar-associated-both questions, and Vivi Nastase and Stan
Szpakowicz for the 600 labeled noun-modifiers. Thanks to Charles Clarke
for the corpus, Stefan B{\"u}ttcher for Wumpus, Guido Minnen, John Carroll,
and Darren Pearce for \emph{morpha} and \emph{morphb}, and Ian Witten and
Eibe Frank for Weka. Thanks to Selmer Bringsjord for inviting me
to contribute to the special issue of \emph{JETAI} on \emph{Test-Based AI}.
Thanks to Joel Martin for comments on an earlier version of this paper.
I am especially thankful to Michael Littman for initiating my interest
in analogies in 2001, by suggesting that a statistical approach might
be able to solve multiple-choice analogy questions.

%
%
\bibliographystyle{named}
\bibliography{turney_jetai09}

\end{document}